\documentclass[review]{elsarticle}

\usepackage{lineno,hyperref}
\modulolinenumbers[5]

\usepackage{graphicx}
\usepackage{algorithm}
\usepackage{algorithmicx}
\usepackage{amsmath,amssymb}
\usepackage{bm}
\usepackage{bbm}
\usepackage{multirow}
\usepackage{caption}
\usepackage{xcolor}
\usepackage{makecell}
\usepackage{bbding}
\usepackage{amsmath,amsfonts}
\usepackage{textcomp}
\usepackage{subfig}
\usepackage{graphicx}

\biboptions{sort&compress}
\newcommand{\etal}{\textit{et al. }}

\journal{}

\begin{document}

\begin{frontmatter}

\title{Deep Progressive Feature Aggregation Network for High Dynamic Range Imaging}

\author{Jun Xiao$^{1}$, Qian Ye$^{2}$, Tianshan Liu$^{1}$, Cong Zhang$^{1}$, Kin-Man Lam$^{1}$}

\address{$^{1}$Department of Electronic and Information Engineering, The Hong Kong Polytechnic University, Hong Kong}
\address{$^{2}$Graduate School of Information Sciences, Tohoku University, Japan}
\fntext[myfootnote]{Email: jun.xiao@connect.polyu.hk (Jun Xiao) and enkmlam@polyu.edu.hk (Kin-Man Lam)}

\begin{abstract}
High dynamic range (HDR) imaging is an important task in image processing that aims to generate well-exposed images in scenes with varying illumination. Although existing multi-exposure fusion methods have achieved impressive results, generating high-quality HDR images in dynamic scenes is still difficult. The primary challenges are ghosting artifacts caused by object motion between low dynamic range images and distorted content in under and overexposed regions. In this paper, we propose a deep progressive feature aggregation network for improving HDR imaging quality in dynamic scenes. To address the issues of object motion, our method implicitly samples high-correspondence features and aggregates them in a coarse-to-fine manner for alignment. In addition, our method adopts a densely connected network structure based on the discrete wavelet transform, which aims to decompose the input features into multiple frequency subbands and adaptively restore corrupted contents. Experiments show that our proposed method can achieve state-of-the-art performance under different scenes, compared to other promising HDR imaging methods. Specifically, the HDR images generated by our method contain cleaner and more detailed content, with fewer distortions, leading to better visual quality.

\end{abstract}
\begin{keyword}
High Dynamic Range Imaging, Image Processing
\end{keyword}

\end{frontmatter}

\section{Introduction}
Most modern imaging systems, e.g., digital cameras, often fail to capture the full range of natural light, because of the hardware, e.g., camera sensors, capacities. These limitations have led to the development of high dynamic range (HDR) imaging technologies, which is essential for generating high-quality images in challenging lighting conditions. The practical applications of HDR imaging are numerous and diverse, such as photography, film, video production, etc. Given its potential for high industrial value, HDR imaging has received significant attention from researchers over the past decades.

In the early stages, researchers attempted to design specialized hardware devices for generating high-quality HDR images \cite{froehlich2014creating,tocci2011versatile}. However, due to their high cost, these devices have not been widely adopted in commercial products. As an alternative, many researchers have turned to methods based on multi-exposure fusion, which take a sequence of low-dynamic range (LDR) images with different exposures and merge them to generate the corresponding HDR images. Although some HDR imaging algorithms can effectively adjust the illumination range based on static images and achieve promising results \cite{yang2018adaptive,liu2020single,eilertsen2017hdr,lee2020learning,tan2021deep,fotiadou2019snapshot}, these methods are primarily suitable for static scenes and less effective when applied to dynamic scenes. Unlike static scenes, HDR imaging in dynamic scenes is more challenging because object motion between input images can result in ghosting artifacts. Additionally, the  under and overexposed regions may contain corrupted content that can lead to distortions. To address these issues, motion-removal-based methods \cite{bogoni2000extending, khan2006ghost, jacobs2008automatic,reinhard2010high,ye2021progressive,gallo2009artifact, grosch2006fast, oh2014robust,yan2017high,lee2014ghost,yan2019attention,pan2020multi} and alignment-based methods \cite{yan2019robust, kalantari2017deep, hafner2014simultaneous, jacobs2008automatic, wu2018deep} have shown their effectiveness in recent years. 

Motion-removal-based methods first detect motion regions and then remove them in the merging stage, so object motion is disregarded in the reconstruction, resulting in reduced ghosting artifacts in the generated images. To accurately detect motion regions, researchers have proposed several methods, including threshold-based methods \cite{grosch2006fast,khan2006ghost,jacobs2008automatic,gallo2009artifact}, gradient-based methods \cite{zhang2011gradient,lee2018multi}, and low-rank and sparsity-based methods \cite{lee2014ghost, oh2014robust, yan2017high}. Firstly, inaccurate motion detection can easily introduce ghosting artifacts, which severely degrade the quality of the generated HDR images. Secondly, when the input LDR images contain large-scale motion, a significant number of pixels are removed in the merging process, leading to information loss and poor performance. Recently, Yan \etal \cite{yan2019attention} proposed a deep learning-based model with the spatial attention mechanism to avoid information loss in moving regions. The model uses a soft suppression method to eliminate unnecessary pixels before the merging stage. However, ghosting artifacts still appear in the generated images, as illustrated in \cite{ye2021progressive}.

Alignment-based methods for HDR imaging involve aligning input LDR images with a reference image, and then merging them to generate the corresponding HDR images. Unlike motion-removal-based methods, alignment-based methods handle object motions using optical flow, which can better preserve information in motion regions. Therefore, they typically achieve better performance. For example, Kalantari \etal \cite{kalantari2017deep} used optical flow \cite{liu2009beyond} to align the input LDR images and then, adopted a deep convolutional neural network (CNN) to merge the aligned LDR images for reconstruction. Their model has demonstrated remarkable performance in producing high-quality HDR images in dynamic scenes. However, the brightness constancy condition required by optical flow is difficult to satisfy in real-world situations \cite{sun2018pwc,yu2016back,teed2020raft}, resulting in inaccurate optical-flow etimation, which further deteriorates the quality of the generated HDR images. Instead of optical flow, Wu \etal \cite{wu2018deep} adopted a homography transformation to globally register LDR images in the pre-processing stage. Then, they used a deep CNN model with the U-Net shape structure to extract multi-scale features for reconstruction. However, this method fails to consider local information, which limits its ability to compensate for corrupted image content in saturated regions. Pu \etal \cite{pu2020robust}, inspired by deformable convolution \cite{zhu2019deformable}, adopted a pyramid, cascaded and deformable (PCD) alignment module \cite{wang2019edvr} to hierarchically align LDR images for reconstruction. Additionally, Liu \etal \cite{liu2021adnet} proposed a dual-branch network for HDR imaging in dynamic scenes, which respectively adopts the spatial attention mechanism and the PCD module in the two branches. However, deformable convolution suffers from unstable training \cite{chan2021understanding}, and its receptive field is restricted by its kernel size, making it difficult to handle large-scale motions.

\begin{figure}[htbp]
    \centering
    \includegraphics[width=\linewidth]{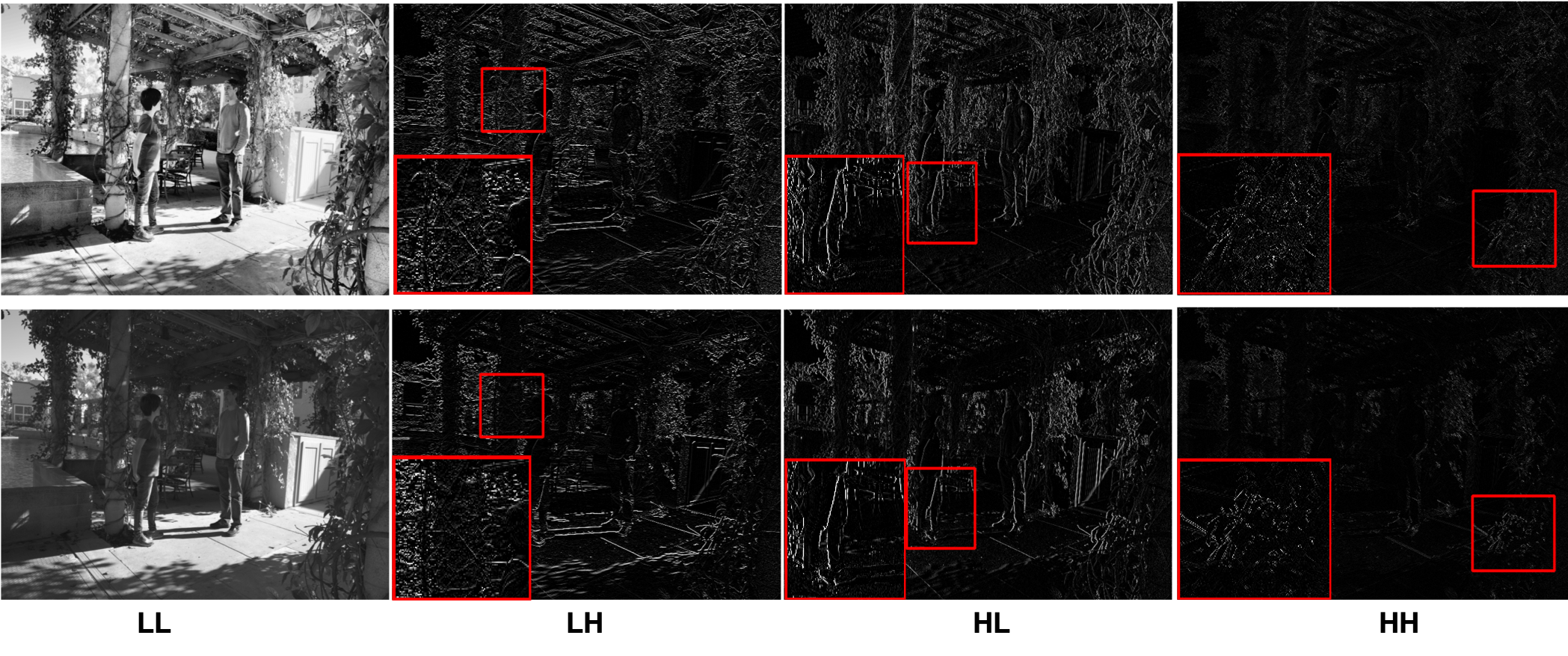}
    \caption{Illustration of an HDR image (the first row) and the corresponding LDR image (the second row) in the wavelet domain. LL, LH, HL, and HH represent the low-low band, low-high band, high-low band, and high-low band, respectively. The region marked by a red box in an image is enlarged and placed at the bottom-left corner.}
    \label{wave_comp}
\end{figure}

In this paper, we propose a deep progressive feature aggregation network for HDR imaging in dynamic scenes. Unlike alignment-based methods, our proposed method does not use optical flow to align input images. Instead, we introduce a cross-scale feature aggregation strategy to implicitly align the input LDR images in a coarse-to-fine manner. In each scale space, our model samples similar features around unaligned pixels and then aggregates them for implicit alignment. These similar features have a high correlation with the unaligned features and, thus, contribute more to alignment. Then, our model progressively fuses the aligned features from coarse to fine scales, which reduces ghosting artifacts caused by small and large motions. To further improve the performance, we propose a densely connected network module based on discrete wavelet transform to effectively compensate for corrupted content in saturated regions. The proposed module decomposes the input features into several non-overlapping frequency subbands and separately restores the corrupted content in these frequency subbands. As shown in Fig.\,\ref{wave_comp}, the low-frequency subband mainly contains coarse image content, while the high-frequency subbands have rich structural information, which is beneficial for HDR imaging. The main contributions of this paper are summarized as follows:
\begin{enumerate}
    \item We propose a novel progressive feature aggregation network for HDR imaging in dynamic scenes. Our method employs intra-scale and inter-scale aggregation schemes to implicitly align LDR images from coarse to fine scales.
    \item We propose a densely connected network based on discrete wavelet transform to effectively compensate for corrupted content in saturated regions.
    \item Experiments demonstrate that our proposed method significantly outperforms other state-of-the-art HDR imaging methods, generating remarkable results with higher visual quality and more structural information.
\end{enumerate}

\section{Related Works}

\subsection{Motion Removal-based Methods}
Motion removal-based methods typically involve two stages: a motion-detection stage and a merging stage. Bogoni \cite{bogoni2000extending} applied a non-parametric model to detect under and overexposed regions before merging LDR images. In \cite{grosch2006fast}, unaligned regions are rejected according to the error map of aligned pixels. Khan \etal \cite{khan2006ghost} proposed an iterative method that implicitly detects moving regions, leading to reduced artifacts. Jacobs \etal \cite{jacobs2008automatic} assumed that moving objects cause large intensity variations in overexposed regions, and proposed a motion-detection method based on measuring variance \cite{reinhard2010high}. Gallo \etal \cite{gallo2009artifact} assumed that irradiance values of the background are linear to the exposure time, and proposed a method for detecting ghosting regions based on the irradiance deviation. All these methods also explode local information in regular patch grids. In contrast, Raman and Chaudhui \cite{raman2011reconstruction} proposed a superpixel-based bottom-up framework to detect ghosting regions, which can effectively handle irregular objects in saturated regions. Heo \etal \cite{heo2010ghost} proposed a coarse-to-fine pipeline that applies a joint probability model to different images to roughly detect motion regions, and then refines these regions with graph cuts. Zhang and Cham \cite{zhang2011gradient} proposed a motion-detection method based on the assumption that gradient values change significantly in motion regions. However, this method struggles to obtain gradient information in saturated regions, which harms its performance. Lee \etal \cite{lee2014ghost} formulated the motion-detection problem as a rank-minimization problem, which considers misalignment errors, moving objects, noise, and nonlinear artifacts as sparse outliers. Oh \etal \cite{oh2014robust} extended the low-rank model in \cite{lee2014ghost} by introducing user control for moving objects under different exposure settings, leading to better results. Yan \etal \cite{yan2019attention} proposed a deep CNN-based model, where the spatial attention mechanism is used to suppress unaligned regions at the feature level, so their model excludes the motion regions in the reconstruction.

Despite their effectiveness, motion removal-based methods have several inherent limitations. First, when there are large-scale motions between LDR images, a large number of pixels may be removed in the merging process, resulting in distorted content. Second, the quality of the generated HDR images is very sensitive to detection accuracy and is not robust to varying illumination.

\subsection{Alignment-based Methods}
Alignment-based methods can be divided into three categories:  flow-based methods, deformable-based methods, and patch-based methods.

\subsubsection{Flow-based methods}

Flow-based methods mainly adopt optical flow to align LDR images and then merge the aligned LDR images to generate the corresponding HDR images. Tomaszewska and Mantiuk \cite{tomaszewska2007image} proposed a global registration method for LDR images using a homography transform before merging them for HDR imaging. However, the condition of brightness constancy is easily violated because of illumination variations. To address this issue,  Zimmer \etal \cite{zimmer2011freehand} proposed an optical flow estimation method in the gradient domain, ensuring consistency of gradients. Although this method is effective, it treats flow estimation and HDR image generation separately, potentially resulting in a sub-optimal solution. To further improve overall performance, Hafner \etal \cite{hafner2014simultaneous} proposed a method for jointly estimating optical flow and reconstructing HDR images. Recently, deep learning-based models have achieved remarkable success in many vision tasks, which inspired Kalantari and Ramamoorthi \cite{kalantari2017deep} to propose a deep learning-based network, called DeepHDR, for HDR imaging. Their method first aligns LDR images using optical flow \cite{liu2009beyond}, and then forwards the aligned LDR images to a deep CNN network for fusion. Instead of using dense motion fields, Wu \etal \cite{wu2018deep} proposed a deep translation-based method that globally registers LDR images with a homography function and then extracts multi-scale features for reconstruction. However, flow-based methods are highly sensitive to varying illumination and rely heavily on the accuracy of optical flow estimation. This sensitivity can result in ghosting artifacts in the generated HDR images.

\subsubsection{Deformable-based methods}

Deep deformable-based methods have recently garnered significant attention in HDR imaging research, following the success of deformable convolution \cite{zhu2019deformable}. Pu \etal \cite{pu2020robust} proposed a progressive alignment and reconstruction approach for HDR imaging, utilizing the PCD module \cite{wang2019edvr}, and achieved promising results. Building on this work, Liu \etal \cite{liu2021adnet} introduced a dual-branch structure comprising the PCD module and spatial attention mechanism, which further improved the overall performance. However, deformable-based methods cannot effectively handle large-scale motions, because the receptive field of the deformable convolutions is restricted by its kernel size. In addition, deformable convolution suffers from unstable training \cite{chan2021understanding}. 

\subsubsection{Patch-based methods}
Compared to flow-based and deformable-based methods, patch-based methods implicitly align the motion regions by aggregating similar image patches. For instance, Sen \etal \cite{sen2012robust} introduced a patch-based HDR imaging method that identifies similar image patches and combines them for reconstruction. Similarly, Yan \etal \cite{yan2020deep} utilized a non-local attention mechanism in a deep CNN model to capture pixel-wise correspondence for image reconstruction. Additionally, Chen \etal \cite{chen2022attention} proposed a hybrid method for HDR imaging that first reduces the motion effect using the spatial attention mechanism and then aggregates the most comparable patch in a coarse-to-fine manner. However, patch-based methods incur high computational complexity, which limits their application in commercial products. To address this challenge, Ye \etal \cite{ye2021progressive} were motivated by RAFT \cite{teed2020raft} and proposed a progressive feature selection approach that implicitly aligns LDR images in the feature space for HDR image reconstruction. Furthermore, Niu \etal \cite{niu2021hdr} proposed a generative adversarial network (GAN)-based model to effectively compensate for corrupted content in saturated regions. However, GAN models are susceptible to model collapse and unstable training \cite{mao2017least,adler2018banach, gulrajani2017improved}, which can lead to artifacts and degrade performance.

Compared with the above methods, our proposed multi-scale sampling and aggregation model has several advantages. Firstly, flow-based methods rely on optical flow for global alignment, whereas our proposed method samples multiple neighboring positions around the unaligned pixels in a coarse-to-fine manner, leading to more accurate and robust alignment. Secondly, our proposed method adaptively aggregates the sampled features that are most similar to the reference features, effectively reducing ghosting artifacts. Thirdly, unlike deformable-based methods which are limited by kernel size, the number of sampled features in our proposed method is flexible, enabling greater adaptability to various image sizes and structures.

\section{The Proposed Methods}
\begin{figure*}
    \centering
    \includegraphics[width=\linewidth]{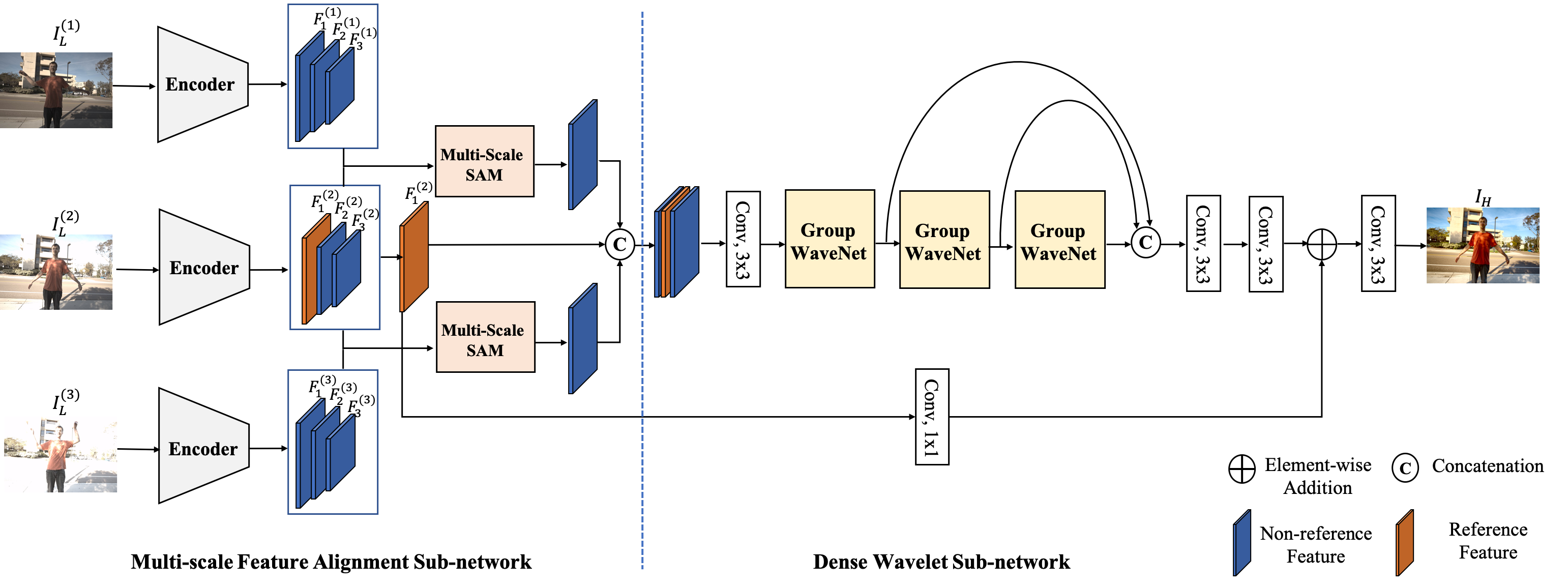}
    \caption{The overall structure of the proposed progressive feature aggregation network, which consists of a multi-scale feature alignment sub-network and a dense wavelet sub-network. $F_{j}^{(i)}$ denotes the features extracted from $X_{i}$ in the $j$-th scale space, where $i,j=1,2,3$.}
    \label{overview}
\end{figure*}

In this section, we present our proposed progressive feature aggregation network, which consists of two crucial components: the multi-scale feature alignment sub-network and the dense wavelet sub-network. An overview of the network architecture is illustrated in Fig.~\ref{overview}. Before delving into the details of these two modules, we first describe the pre-processing techniques that we have employed for the input LDR images. Then, we provide a comprehensive explanation of the multi-scale correspondence alignment sub-network and the dense wavelet sub-network. The former computes the high-correspondence features to implicitly align motion regions in a coarse-to-fine manner. The latter utilizes discrete wavelet transform to decompose the input features into several non-overlapping frequency subbands for compensating corrupted content in saturated regions. Finally, we discuss the loss function used for training our proposed model.

\begin{figure*}
    \centering
    \includegraphics[width=\linewidth]{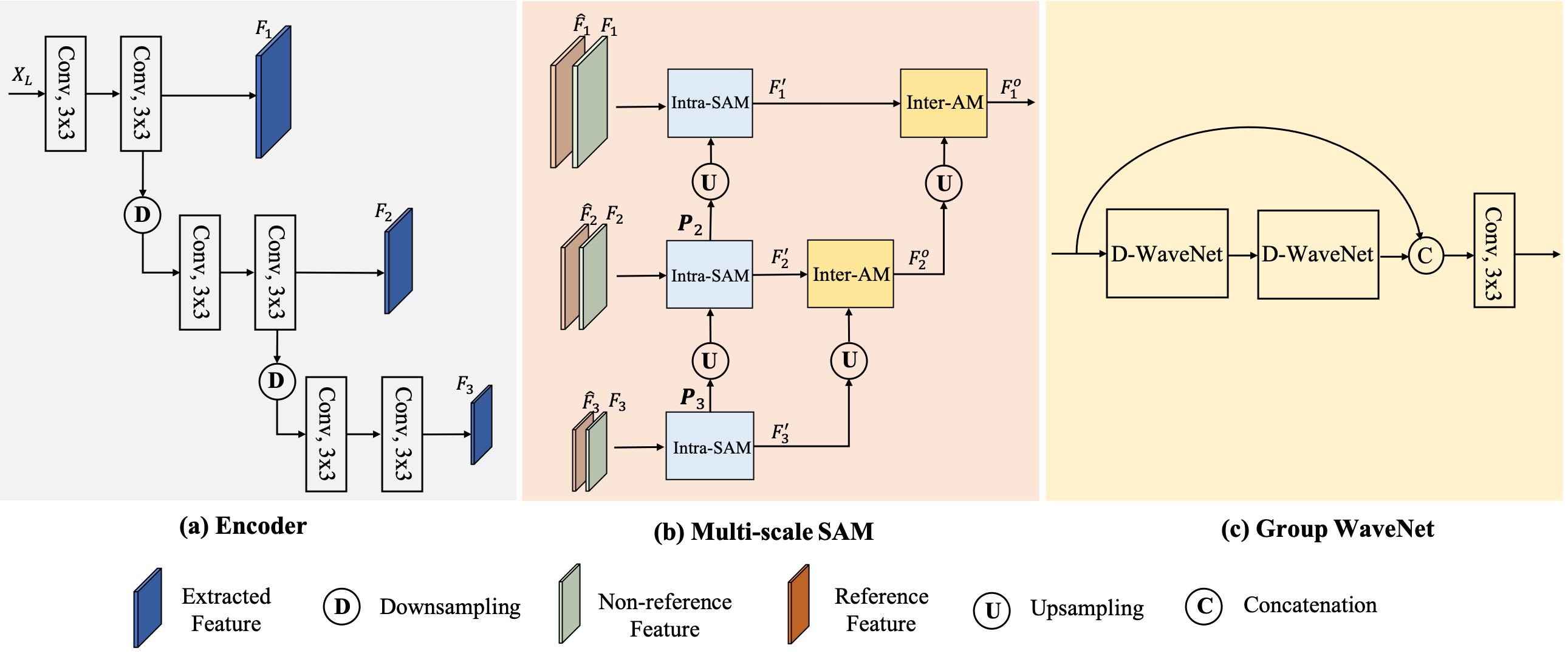}
    \caption{The overall structure of (a) the encoder, (b) Multi-Scale SAM, and (c) Group WaveNet. In the encoder, $X_{L}$ denotes the input 6-channel tensor, and $F_{i}$ represents the extracted features of different scales, for $i=1,2,3$. In the Multi-scale SAM, $\hat{F}_{i}$ and $F_{i}$ denote the reference and non-reference features, respectively, where $i=1,2,3$. }
    \label{compents}
\end{figure*}

\subsection{The Pre-processing Stage}
Given a sequence of LDR images $\{I_{L}^{(1)},I_{L}^{(2)},\cdots, I_{L}^{{(N)}}\}$ with $N$ different exposure times, our model aims to reconstruct a clean and ghost-free HDR image $I_{H}^{(r)}$, which is aligned to the reference image $I_{L}^{(r)}$, where $r\in \{1,2,\cdots, N\}$. To simplify the problem, we consider a sequence of three LDR images with different exposure conditions: low, medium, and high, denoted as $\{I_{L}^{(1)}, I_{L}^{(2)}, I_{L}^{(3)}\}$. We take the LDR image $I_{L}^{(2)}$, captured with a medium exposure time, as our reference. As HDR imaging involves manipulating illumination values, we first linearize the input LDR images with gamma correction, where $\gamma>1$ is a hyper-parameter. Specifically, the linearized LDR image $\hat{I}^{(i)}_{L}$ of the $i$-th input LDR image $I_{L}^{(i)}$ is computed as follows:
\begin{equation}
    \hat{I}^{(i)}_{L}=\frac{\left(I_{L}^{(i)}\right)^{\gamma}}{t_{i}}, \quad i=1,2,3,
\end{equation}
where $t_{i}$ denotes the exposure time of the LDR image $I_{L}^{(i)}$. We set $\gamma$ to $2.2$, as suggested in \cite{kalantari2017deep}. Then, we concatenate each LDR image $I_{L}^{(i)}$ with its corresponding linearized image $\hat{I}^{(i)}_{L}$ to form a 6-channel tensor $X_{i}=[I_{L}^{(i)}, \hat{I}^{(i)}_{L}]$, for $i=1,2,3$, which is the input to the model.

Given a number of HDR-LDR image pairs, our model attempts to learn a highly nonlinear illumination mapping from the LDR domain to the HDR domain. The estimated HDR image $\hat{I}_{H}$ is generated as follows:
\begin{equation}
    \hat{I}_{H} = f_{\theta}(X_{1}, X_{2}, X_{3}),
\end{equation}
where $f_{\theta}(\cdot)$ denotes the proposed model with parameters $\theta$.

\subsection{The Multi-scale Feature Alignment Sub-network}

The multi-scale feature alignment sub-network plays a crucial role in our proposed model, which implicitly aligns the input LDR images from coarse to fine levels. The overall architecture of this sub-network is illustrated in Fig.\,\ref{overview}, which comprises two components: an encoder and a multi-scale sampling and aggregation module (Multi-scale SAM).

\subsubsection{Encoder} The encoder takes $\{X_{1}, X_{2}, X_{3}\}$ as input and extracts the corresponding multi-scale features. As shown in Fig.\,\ref{compents}, the encoder in our method contains three scale spaces. In each scale space, two convolutional layers with a kernel size of $3\times 3$ are used to generate features. The downsampling operator is a convolutional layer with a stride of $2$. In our method, the kernel weights of the encoders are shared among the three inputs to avoid increasing the model parameters. The generated multi-scale features are denoted as $\bm{F}_{i}=\{F_{i}^{(1)}, \cdots, F_{i}^{(S)}\}$, for $i=1,2,3$, where $S$ is the number of scales.  Then, these multi-scale features are forwarded to multi-scale SAM.

\subsubsection{Multi-scale SAM} 
The responsibility of Multi-scale SAM is to implicitly align the LDR images in feature spaces, and the overall structure of this module is shown in Fig.\,\ref{compents}(b). As observed, multi-scale SAM involves two important parts, namely, the intra-scale sampling and aggregation module (Intra-SAM) and the inter-scale aggregation module (Inter-AM).

\begin{figure}[htbp]
    \centering
    \includegraphics[width=\linewidth]{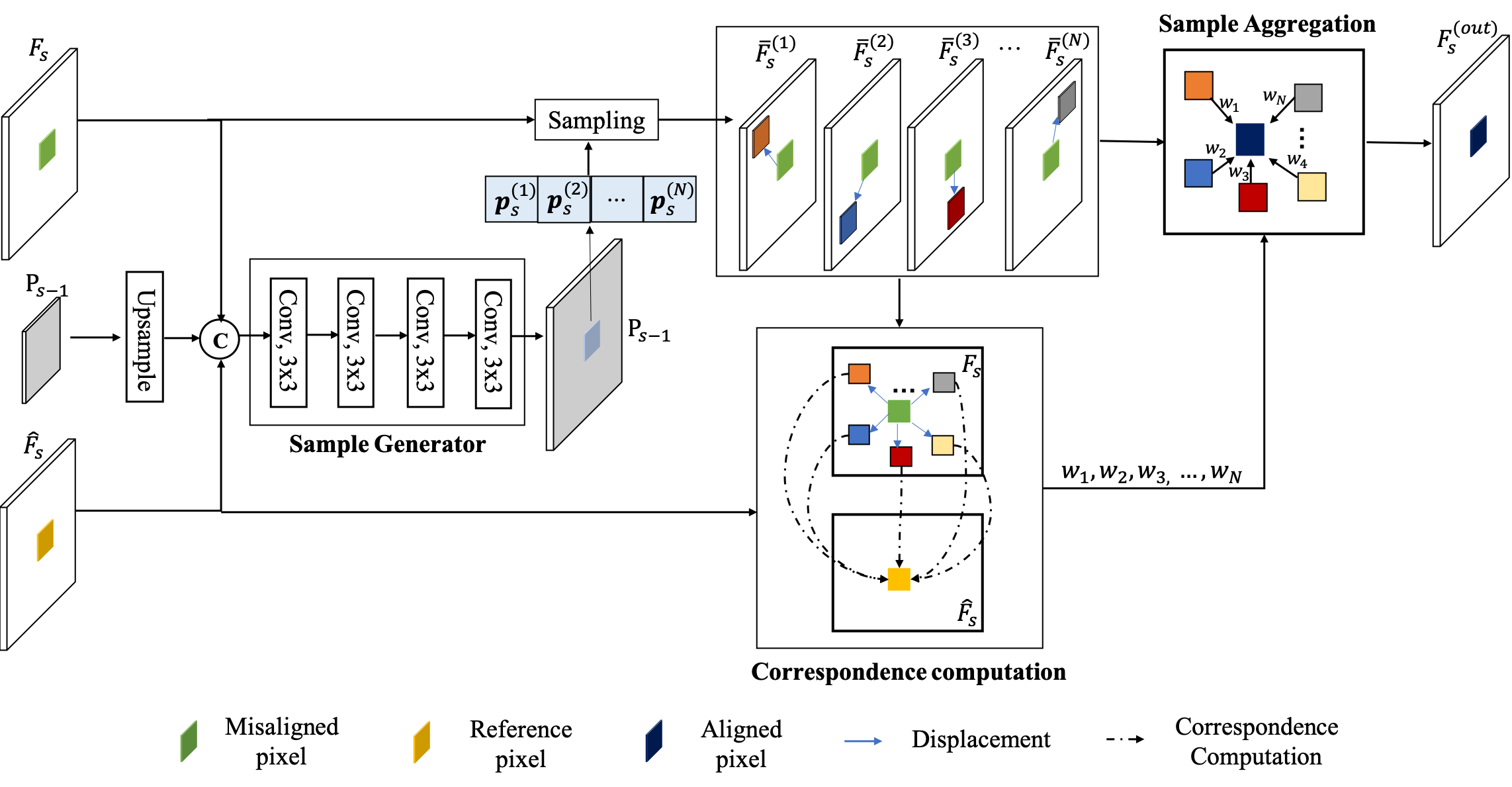}
    \caption{The overall structure of the sampling and aggregation module (SAM). $F_{s}$ and $\hat{F}_{s}$ denote the non-reference and reference features of the $s$-th scale space. $P_{s}$ and $P_{s-1}$ are the sets of sampling positions. $\bm{p}_{s}^{(i)}$ denotes the $i$-th sampling position in the $s$-th scale space, where $i=1,\cdots N$ and the $N$ is the number of samples. $\bar{F}^{(i)}_{s}$ denotes the $i$-th sample feature in the $s$-th scale space. $\{w_{i},\cdots, w_{N}\}$ are $N$ aggregation weights. $F^{(\text{out})}_{s}$ is the output feature associated with $F_{s}$. }
   \label{SAM}
\end{figure}

\textbf{Intra-SAM}. Intra-SAM takes features from each scale space as the input for alignment. Its primary function is to align pixels in the non-reference features by sampling neighboring features, and then, adaptively aggregating the features according to their correspondence with the reference features. As illustrated in Fig.\,\ref{SAM}, the proposed Intra-SAM module comprises three parts: sample generator, correspondence computation, and sample aggregation.

Given two input features in the $s$-th scale space, denoted as $F_{s}$ and $\hat{F}_{s}$, which are the features extracted from the non-reference image and the reference image, respectively, the sample generator takes these two features, as well as the sampling map $P_{s-1}$ from the previous scale space, as input, to generate the sampling map of the $s$-th scale space $P_{s}$. The sampling map $P_{s-1}$ contains the coarse sampling information of the $(s-1)$-st scale space, which can be used as prior information to guide the generation of the sampling map or matrix of the $s$-th scale space. In our method, the sample generator is a small convolutional network with four convolutional layers, and the sampling map $P_{s}$ is computed, as follows:
\begin{equation}
    P_{s} = g(F_{s},\hat{F}_{s}, \hat{P}_{s-1}),
\end{equation}
where $g(\cdot)$ denotes the sampling generator and $P_{s}=[\bm{p}_{s}^{(1)},\cdots, \bm{p}_{s}^{(N)}]$ represents a set contains $N$ sampling matrices in the $s$-th scale space, where the dimension of the sampling matrix $\bm{p}_{s}^{(i)}$ is $H\times W\times 2$, for $i=1,\cdots, N$. $H$ and $W$ represent the height and width of the input features, respectively. These sampling matrices provide the location information of the corresponding sampled feature. Specifically, for the $s$-th scale space, the element of the $i$-th sampling matrix at the position $(m, n)$ is a two-dimensional vector, denoted as $\bm{p}^{(i)}_{s,m,n} = [x_{s,m,n}^{(i)}, y_{s,m,n}^{(i)}]$. This vector contains displacement information between the sampled feature and the reference feature in the horizontal and vertical directions. Based on the sampling matrices, the sampled feature of pixel position $(m,n)$ is obtained as follows:
\begin{equation}
    \bar{F}^{(i)}_{s}(m,n) = F_{s}(m+x_{s,m,n}^{(i)}, n+y_{s,m,n}^{(i)}),
\end{equation}
where $\bar{F}^{(i)}_{s}(m,n)$ denotes the $i$-th sampled feature corresponding to the unaligned pixel at the position $(m,n)$ of the $s$-th scale space, for $i=1,\cdots, N$. For fractional positions (i.e., $m+x_{s,m,n}^{(i)}, n+y_{s,m,n}^{(i)} \notin\mathbb{Z}$), bilinear interpolation is used to compute the positions. In the $s$-th scale space, intra-SAM samples the neighboring features for all unaligned pixels according to the location information provided by the sampling map $P_{s}$. 

However, not all the sampled features are beneficial for reconstructing HDR images, especially in saturated regions. To address this issue, we propose an adaptive aggregation method that performs over the sampled features. Specifically, based on dot product, we first calculate the correspondence between the sampled features and the reference features at each pixel position. Then, we compute the correspondence weights for aggregation. For position $(m,n)$, the weight is computed as follows: 
\begin{equation}
        w^{(i)}_{m,n} =\frac{\exp(\bar{F}^{(i)}_{s}(m,n) \cdot \hat{F}_{s}(m,n))}{\sum_{j=1}^{N}\exp(\bar{F}^{(j)}_{s}(m,n) \cdot \hat{F}_{s}(m,n))}, \label{correspondence}
\end{equation}
where $\cdot$ denotes the dot product, and $0\leq w^{(i)}_{m,n}\leq 1$, for $i=1,\cdots, N$.

Next, the non-reference features are aligned by aggregating the sampled features according to the computed weights, and the corresponding aligned feature $\hat{F}_{s}(m,n)$ is computed as follows:

\begin{equation}
        \hat{F}_{s}(m,n) = w^{(1)}_{m,n} \bar{F}^{(1)}_{s}(m,n) + \cdots + w^{(N)}_{m,n} \bar{F}^{(N)}_{s}(m,n).
\end{equation}
As observed, aligned features are generated by a linear combination of the sampled features. Intuitively, the sampled features having higher correspondence mean that they are more similar to the reference features, so they can contribute more to alignment and should be given more weight in the aggregation.This can effectively alleviate undesirable effects caused by dissimilar features in aggregation and avoid ghosting artifacts.

\begin{figure}
    \centering
    \includegraphics[width=0.7\linewidth]{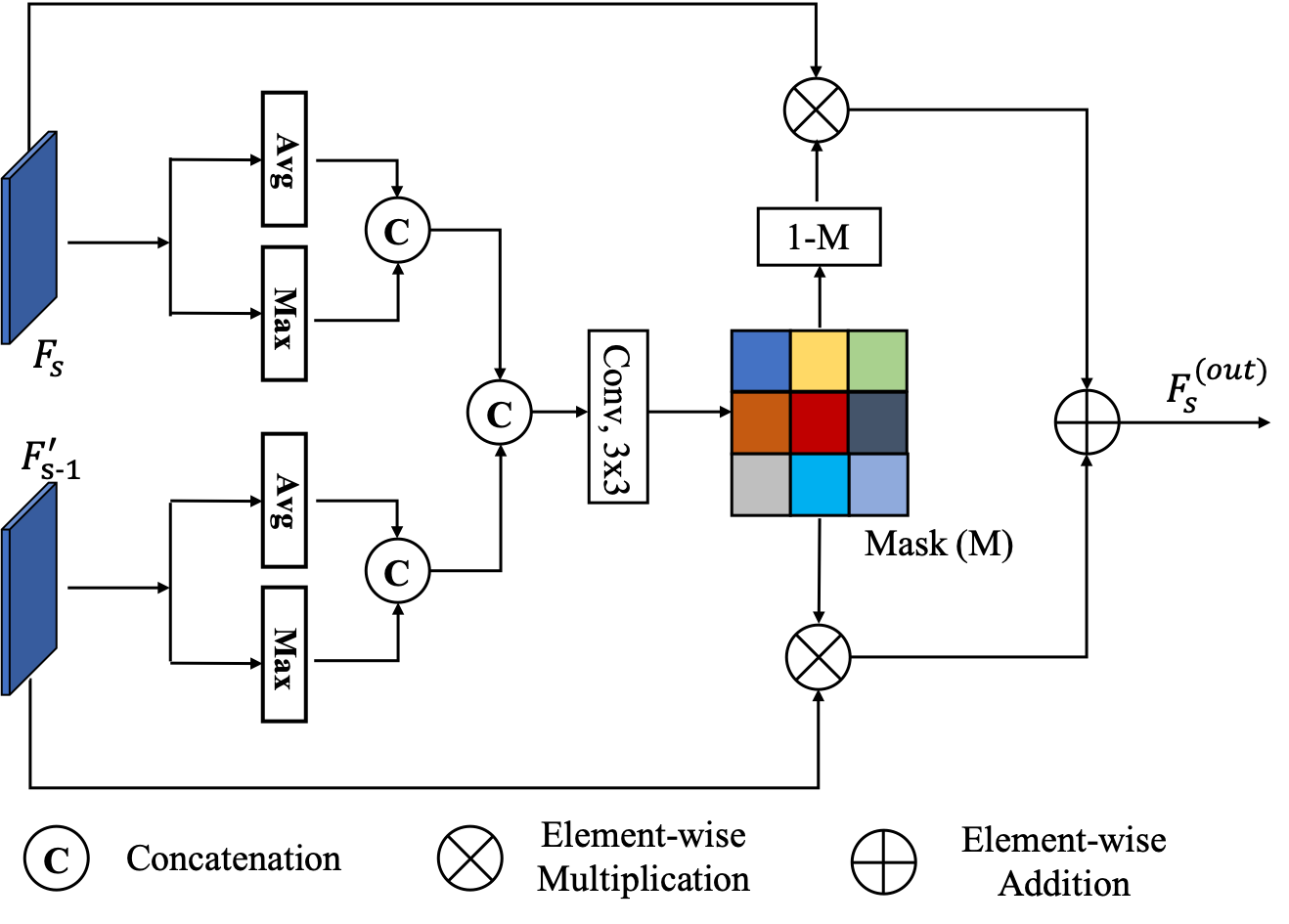}
    \caption{The overall structure of the inter-scale aggregation module (Inter-AM). $F_{s}$ and $F^{\prime}_{s-1}$ denote the input from the $s$-th and $(s-1)$-th scale spaces, respectively. ``Avg" and `` Max" represent average pooling and max pooling, respectively. $F_{s}^{\text{out}}$ is the output feature of the $s$-th scale space.}
    \label{fig:ISAM}
\end{figure}

\textbf{Inter-AM}. To tackle the challenges of small and large-scale motion in dynamic scenes present in LDR images, the proposed Inter-AM aims to progressively aggregate the features generated by Intra-SAM in different scale spaces. The overall network architecture of Inter-AM is illustrated in Fig.\,\ref{fig:ISAM}.

Given two output features $\hat{F}_{s}$ and $\hat{F}^{\prime}_{s-1}$ from the $s$-th and $(s-1)$-st scales, average pooling and max pooling are applied to these two features to extract local features. Then, the local features are concatenated along the channel dimension. A convolutional layer is applied to the concatenated features to generate a fusion mask $M$. Each element of the fusion mask $M$ is in $[0, 1]$. The two input features from adjacent scales are adaptively aggregated based on this fusion mask to produce the corresponding multi-scale feature $F_{s}^{(\text{out})}$, which is computed as follows:
\begin{equation}
    F_{s}^{(\text{out})} = (1-M)\odot F_{s} + M\odot F^{\prime}_{s-1},
\end{equation}
where $\odot$ denotes the Hadamard product. The two input features are spatially fused by linear combination, based on the computed fusion mask $M$. The coarse-scale features contain more contextual information, while the finer-scale features have more detailed information. The fusion mask performs feature selection between the two input features from adjacent scales in a soft manner. Benefiting from this multi-scale structure, the negative impacts of small and large-scale motions in LDR images are eliminated in a coarse-to-finer manner.

\subsection{Dense Wavelet Sub-network}
\subsubsection{Network Structure}
As illustrated in Fig.\,\ref{overview}, the dense wavelet sub-network is responsible for merging aligned features and generating corresponding HDR images. Specifically, the dense wavelet sub-network first concatenates the aligned features (i.e., $F^{1\rightarrow 2}_{1}$ and $F^{3\rightarrow 2}_{1}$) and the reference feature $F^{2}_{1}$ along the channel dimension, where $F^{i\rightarrow 2}_{1}$ denotes the features generated by aligning $F^{i}_{1}$ with the reference feature $F^{2}_{1}$, where $i=1$ or $3$. Then, a convolutional layer with a kernel size of $3\times 3$ is applied to compress and fuse the concatenated features. The resulting fused feature is forwarded to three group-wavelet modules (Group WaveNets) for feature extraction in the wavelet domain. As shown in Fig.\,\ref{compents}(c), a Group WaveNet, which is composed of two small, cascaded dense wavelet networks (D-WaveNets), plays a significant role in feature extraction and HDR image reconstruction. In our method, the output of the second D-WaveNet is concatenated with the input, followed by a convolutional layer for feature fusion. Furthermore, the output of the three Group WaveNets are concatenated, followed by two convolutional layers for fusion. Therefore, the proposed dense wavelet sub-network has a structure with local-global dense connections. This can significantly increase information interaction and fully utilize the features from different layers. At the output of the network, the estimated HDR $\hat{I}_{\text{H}}$ image is computed, as follows:
\begin{equation}
    \hat{I}_{\text{H}}= F^{\prime} + \text{Conv}(F^{2}_{1}),
\end{equation}
where $F^{\prime}$ denotes the fused feature generated by the two convolutional layers and $\text{Conv}(\cdot)$ represents a convolutional layer with a kernel size of one.   

\subsubsection{Dense Wavelet Network (D-WaveNet)}
\begin{figure}
    \centering
    \includegraphics[width=0.9\linewidth]{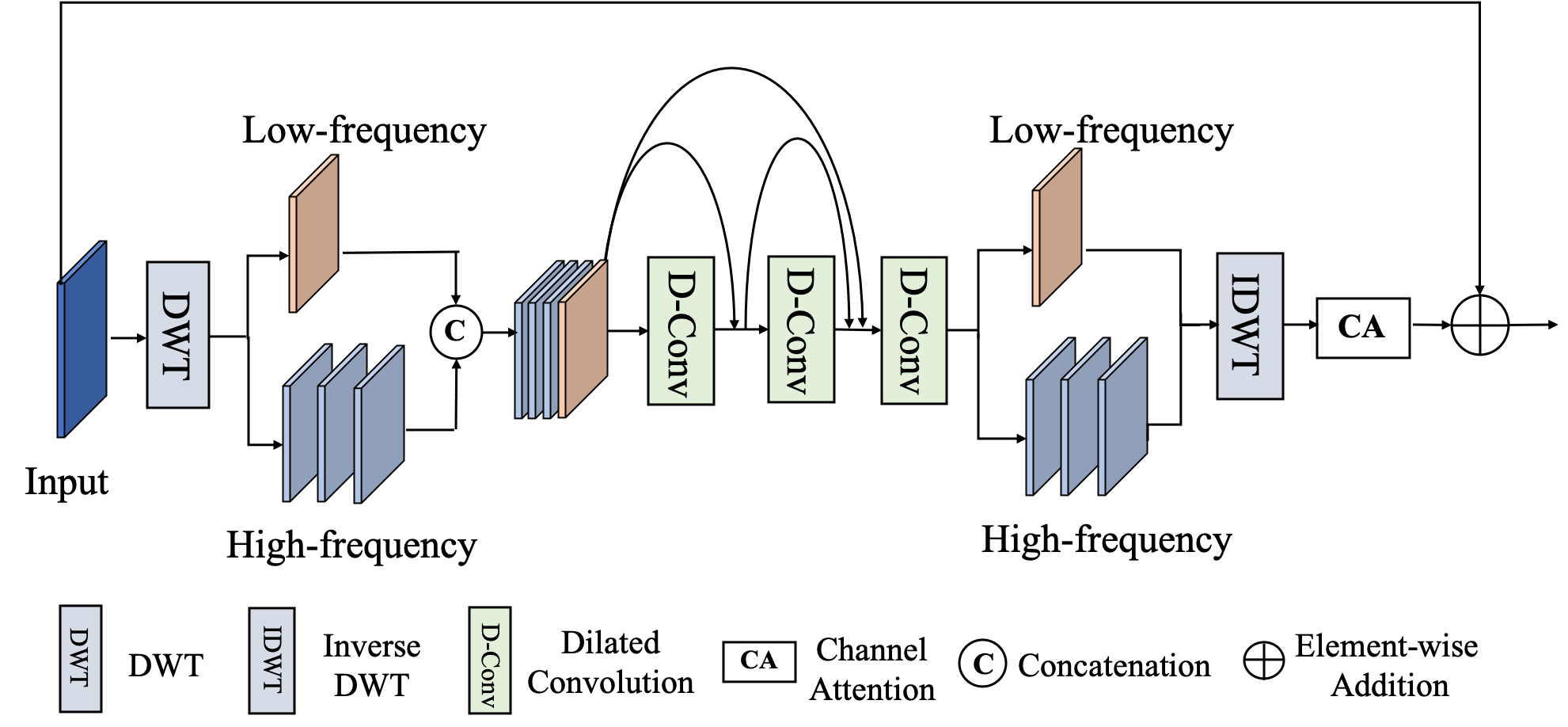}
    \caption{The overall structure of D-WaveNet. The dilated rates of the three dilated convolutional layers are set to 1, 2, and 3, respectively.}
    \label{fig:wavenet}
\end{figure}

D-WaveNet, based on discrete wavelet transform (DWT), decomposes input features into several non-overlapping frequency subbands for feature extraction. The overall structure of D-WaveNet is illustrated in Fig.\,\ref{fig:wavenet}. Given an input feature map $F$, DWT is applied to $F$ along the channel dimension, and then four frequency subbands are obtained, i.e., one low-frequency subband and three high-frequency subbands. These four frequency subbands are computed as follows:
\begin{equation}
    W_{LL} = \mathcal{L} F\mathcal{L}^{T}, W_{LH} = \mathcal{H} F\mathcal{L}^{T}
\end{equation}
\begin{equation}
    W_{HL} = \mathcal{L} F\mathcal{H}^{T}, W_{HH} = \mathcal{H} F\mathcal{H}^{T}
\end{equation}

where $W_{LL}$ denotes the low-frequency subband and $\{W_{LH}, W_{HL}, W_{HH}\}$ represent the high-frequency sub-bands in the horizontal, vertical, and diagonal directions, respectively. $\mathcal{L}$ denotes the matrix containing all the low-frequency filters $\{\ell_{i}\}_{i\in\mathbb{Z}}$ and $\mathcal{H}$ is the matrix containing all the high-frequency filters $\{h_{i}\}_{i\in\mathcal{Z}}$ for decomposition, where
\begin{equation}
    \mathcal{L} = \begin{pmatrix}
        \cdots &  \cdots & \cdots & & & & \\
         \cdots&  \ell_{-1} & \ell_{0} & \ell_{1} & \cdots & \cdots & \\
          & & \cdots&  \ell_{-1} & \ell_{0} & \ell_{1} & \cdots &  \\
          & & &   &  &  & \cdots & \cdots  
    \end{pmatrix},
\end{equation}
and
\begin{equation}
    \mathcal{H} = \begin{pmatrix}
        \cdots &  \cdots & \cdots & & & & \\
         \cdots&  h_{-1} & h_{0} & h_{1} & \cdots & \cdots & \\
          & & \cdots&  h_{-1} & h_{0} & h_{1} & \cdots &  \\
          & & &   &  &  & \cdots & \cdots  
    \end{pmatrix}.
\end{equation}
In our method, Haar wavelets are adopted for feature decomposition, because Haar wavelets have orthogonal and biorthogonal properties. These two properties guarantee that the input features can be perfectly reconstructed, without concerns about introducing distortions. After decomposition, all frequency subbands are concatenated along the channel dimension and forwarded to three dilated convolutional layers with dense connections for feature extraction over the wavelet domain. The dilated rate of these three convolutional layers is set to 1, 2, and 3, respectively, in our method.

At the output of D-WaveNet, the extracted features are transferred from the wavelet domain to the spatial domain with the inverse DWT (IDWT), which is computed, as follows:
\begin{equation}
    \hat{F} = \mathcal{L}^{T} \hat{W}_{LL}\mathcal{L} + \mathcal{H}^{T} \hat{W}_{LH}\mathcal{L} + \mathcal{L}^{T} \hat{W}_{HL}\mathcal{H} + \mathcal{H}^{T} \hat{W}_{HH}\mathcal{H},
\end{equation}
where $\{\hat{W}_{LL}, \hat{W}_{LH}, \hat{W}_{HL}, \hat{W}_{HH}\}$ denote four output wavelet subbands, and $\hat{F}$ is the corresponding output feature in the spatial domain. Then, we use a residual connection and the channel attention mechanism \cite{hu2018squeeze} to generate the output feature $F_{\text{out}}$ as follows:
\begin{equation}
    F_{\text{out}}=\text{CA}(\hat{F})+F,
\end{equation}
where $\text{CA}(\cdot)$ is the channel attention mechanism, which scales features along the channel dimension.

\subsection{Loss Function}
As HDR images are typically displayed on screens after tone mapping, we calculate the loss in tone-mapped images. Although many effective tone-mapping methods have been proposed in the past, they are typically not differentiable and, therefore, cannot be used as a loss function for training deep neural networks. In our method, we apply the $\mu$-law function to compress the output of our network, because the $\mu$-law function  is a widely used range compressor in audio processing and is differentiable. Specifically, given an image $I$ in the linear domain, the corresponding tone-mapped image is computed, as follows:
\begin{equation}
    T(I)=\frac{\log(1+\mu I)}{\log (1+\mu)},
\end{equation}
where $\mu$ is a hyper-parameter defining the extent of compression, set to $5000$ in our method. 
We use the $L_{1}$ norm to compute the distance between the output  $\hat{I}_{H}$ and the ground-truth image $I_{H}$ after tone mapping. Formally, the loss function is defined as follows:
\begin{equation}
    L(\hat{I}_{H}, I_{H})=\Arrowvert T(\hat{I}_{H})-T(I_{H})\Arrowvert_{1}.
\end{equation}
In our method, the range of the output  is restricted to $[0,1]$ by using the sigmoid function.

\subsubsection{Implementation details}
To accommodate the large size of the training images, we partitioned the input images into patches of size $200\times 200$ pixels for training. To augment the training dataset, all training samples are randomly flipped vertically or horizontally, as well as rotated by $90^{\circ}$, $180^{\circ}$, or $270^{\circ}$. We adopt the AdamW algorithm,  with $\beta_{1}=0.9$ and $\beta_{2}=0.999$, to adaptively update the parameters of the proposed model during training. The batch size is set to $16$, and the initial learning rate is set to $2.0\times 10^{-4}$. To adjust the learning rate, we employ the cosine annealing strategy, which gradually reduces the learning rate to $1.0\times 10^{-6}$. Our proposed model was implemented using the Pytorch framework, with two Nvidia 3090 GPUs, and we trained the model for a total of 300 epochs. The training process took approximately two days to complete. 


\begin{table}[htbp]
\caption{The average PSNR-$\mu$, SSIM-$\mu$, PSNR-L, SSIM-L, and HDR-VDP-2 of different methods on the Kalantari dataset. The best results are highlighted in bold.}
\begin{tabular}{|c|c|c|c|c|c|}
\hline
Methods & PSNR-$\mu$ & PSNR-L & SSIM-$\mu$ & SSIM-L & HDR-VDP-2 \\ \hline
Sen \cite{sen2012robust} & 40.9689 & 38.3425 & 0.9859 & 0.9764 & 60.3463 \\ \hline
DeepHDR \cite{kalantari2017deep} & 42.7177 & 41.2200 & 0.9889 & 0.9829 & 61.3139  \\ \hline
Wu \cite{wu2018deep} & 41.9977 & 41.6593 & 0.9878 & 0.9860 & 61.7981 \\ \hline
AHDRNet \cite{yan2019attention} & 43.7013 & 41.1782 & 0.9905 & 0.9857 & 62.0521 \\ \hline
PANet \cite{pu2020robust} & 43.8487 & 41.6452 & 0.9906 & 0.9870 & 62.5495 \\ \hline
PSFNet \cite{ye2021progressive} & 44.0613 & 41.5736 & 0.9907 & 0.9867 & 63.1550 \\ \hline
PFANet (Ours) & \textbf{44.3847} & \textbf{42.1704} & \textbf{0.9887} & \textbf{0.9887} & \textbf{64.5038} \\ \hline
\end{tabular}
\label{exp_one}
\end{table}

\begin{figure}[!t]
    \centering
\subfloat{\includegraphics[width=0.9\linewidth]{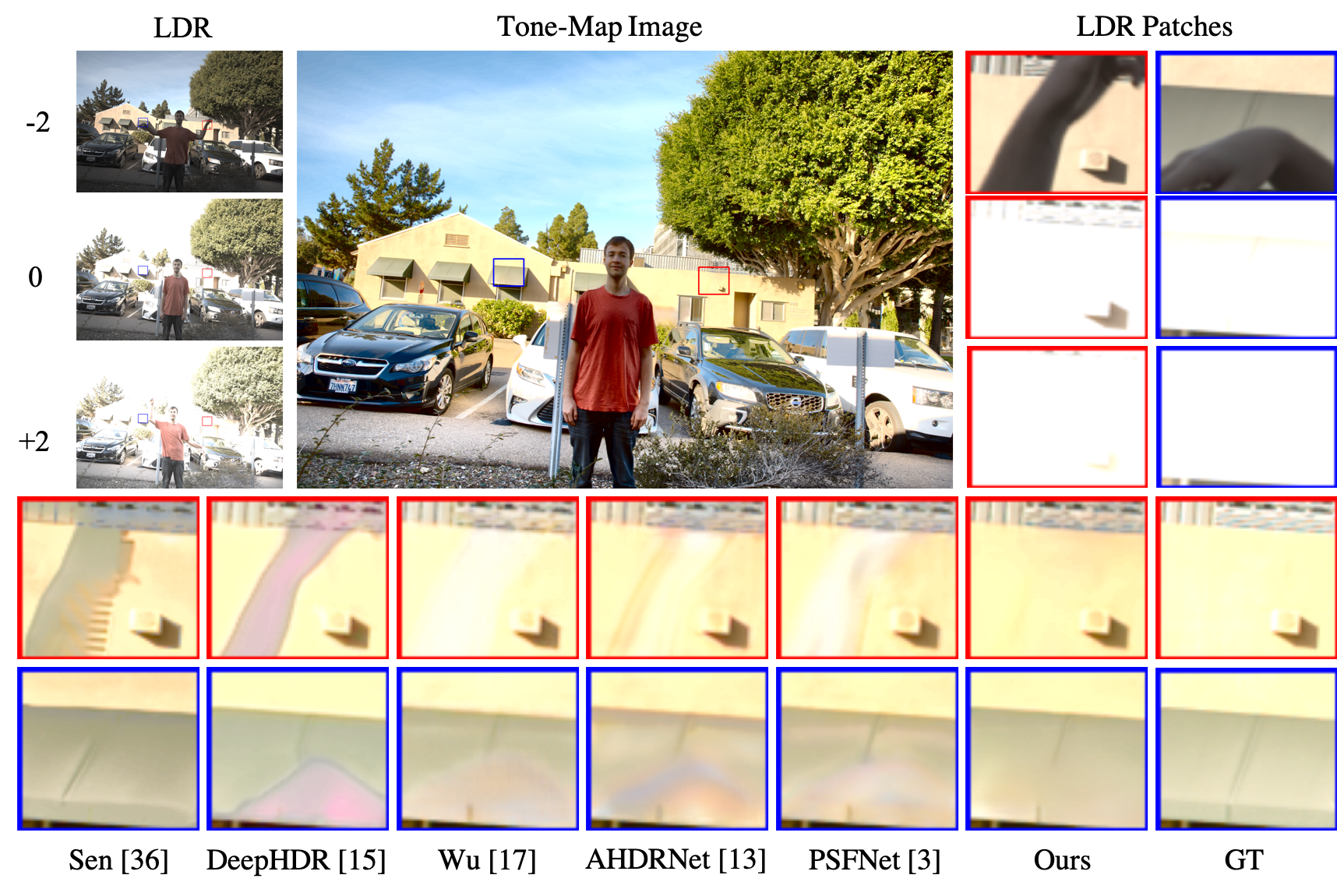}
}\\
\subfloat{\includegraphics[width=0.9\linewidth]{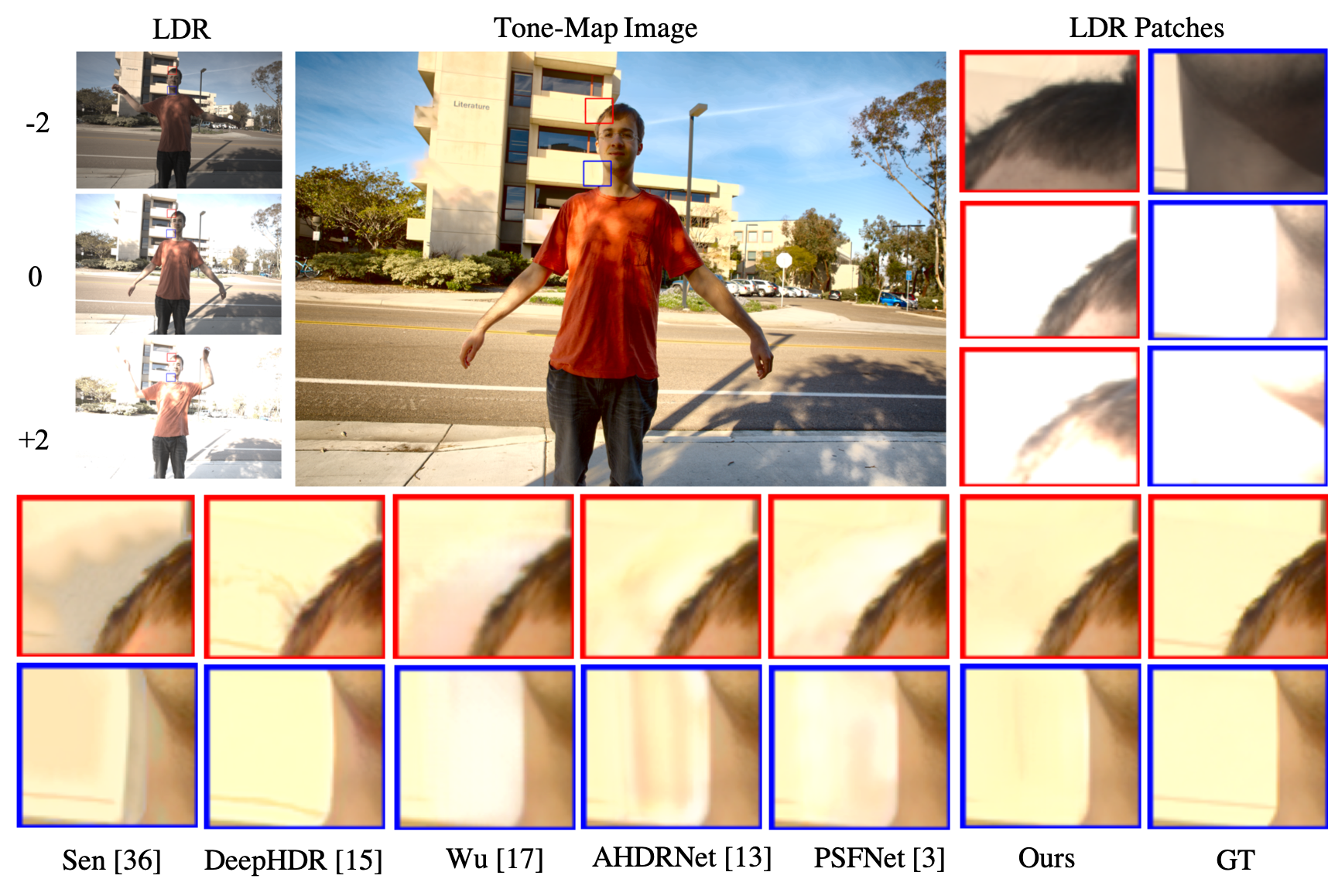}
}
\caption{Visual comparison of different HDR imaging methods. ``-2", ``0", and ``+2" denote three different exposure settings. In the experiment, ``0" represents the reference image.}
    \label{vis_1}
\end{figure}

\subsection{Experiments on the Kalantari Dataset}
In this experiment, we compare PFANet with other state-of-the-art HDR methods, including Sen \cite{sen2012robust}, DeepHDR \cite{kalantari2017deep}, Wu \cite{wu2018deep}, AHDRNet \cite{yan2019attention}, PANet \cite{pu2020robust}, and PSFNet \cite{ye2021progressive}. We utilized the open-source code provided by the respective authors for implementation, with the exception of PANet, which has no publicly available code. Instead, we referred to the results presented in the original paper. The average PSNR-$\mu$, PSNR-L, SSIM-$\mu$, SSIM-L, and HDR-VDP-2 of different HDR methods on the Kalantari dataset are tabulated in Table \ref{exp_one}. The results show that PFANet outperforms all other HDR methods in all evaluation metrics. Compared to the second-best method (PSFNet), PFANet achieves a gain of 0.3dB and 1.3488, in terms of PSNR-$\mu$ and HDR-VDP-2, respectively. These results indicate that our proposed method can better generate high-quality HDR images in both tone-mapped and linear domains than other HDR methods.

To further evaluate the performance of the compared methods, we selected two images from the Kalantari dataset and visually compare the results, as shown in Fig.\,\ref{vis_1}. We zoom into two local regions marked by red and blue rectangles, which contain moving objects, for better comparison. As observed, the results generated by DeepHDR, Wu, AHDRNet, and PSFNet exhibit undesirable ghosting artifacts and distorted contents in the motion and overexposed regions. Although Sen is successful in restoring the corrupted content in saturated regions, it fails to handle motion regions and produces severe ghosting artifacts. These results reveal that the compared methods struggle to simultaneously compensate for corrupted content in saturated regions and overcome object motions. In contrast, PFANet effectively alleviates ghosting artifacts caused by object motions and adequately compensates for the corrupted regions in under and overexposed regions. As a result, the HDR images generated by PFANet contain less distorted image content, leading to the best visual quality.

\begin{figure*}[htbp]
    \centering
\subfloat{\includegraphics[width=0.9\linewidth]{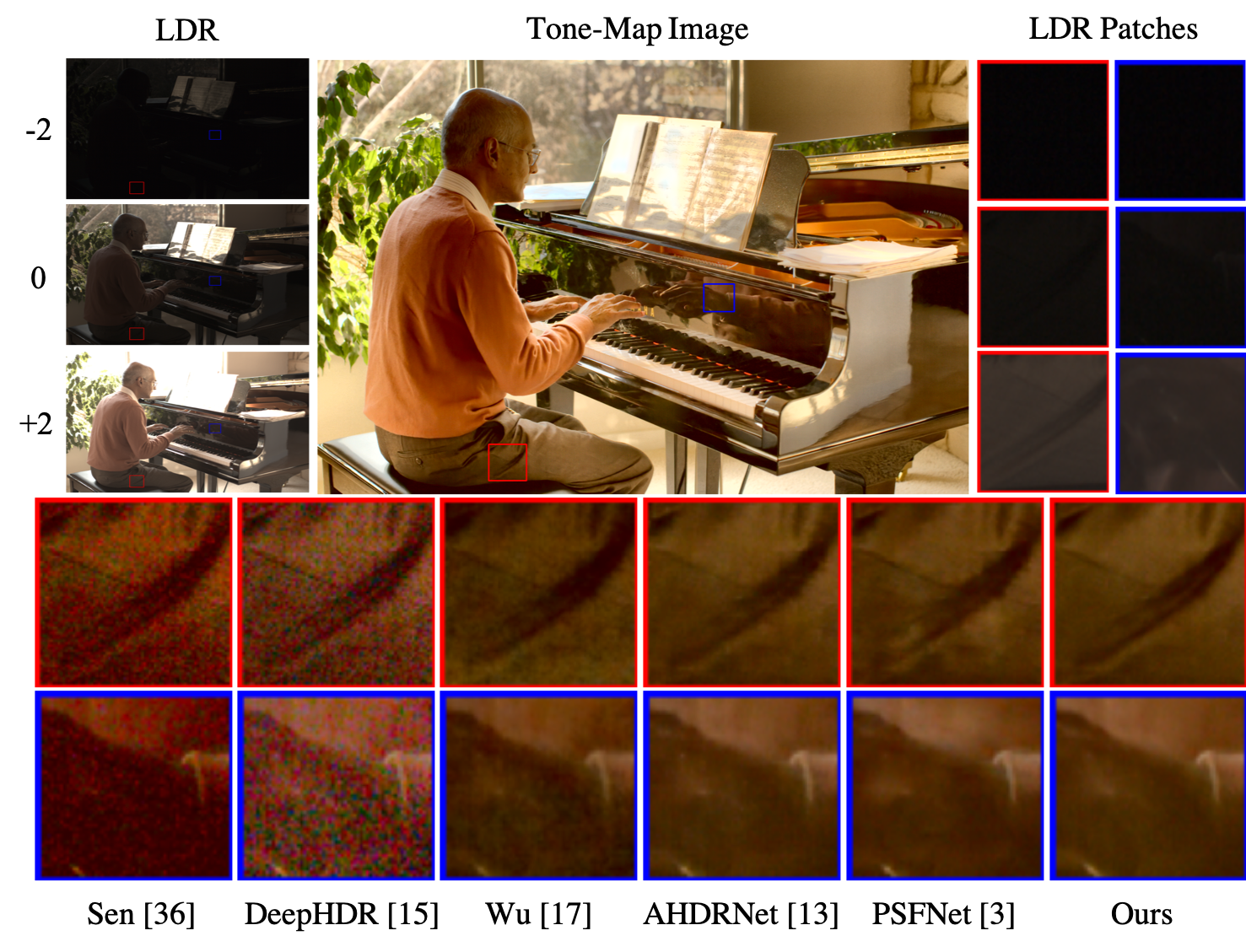}
}\\
\subfloat{\includegraphics[width=0.9\linewidth]{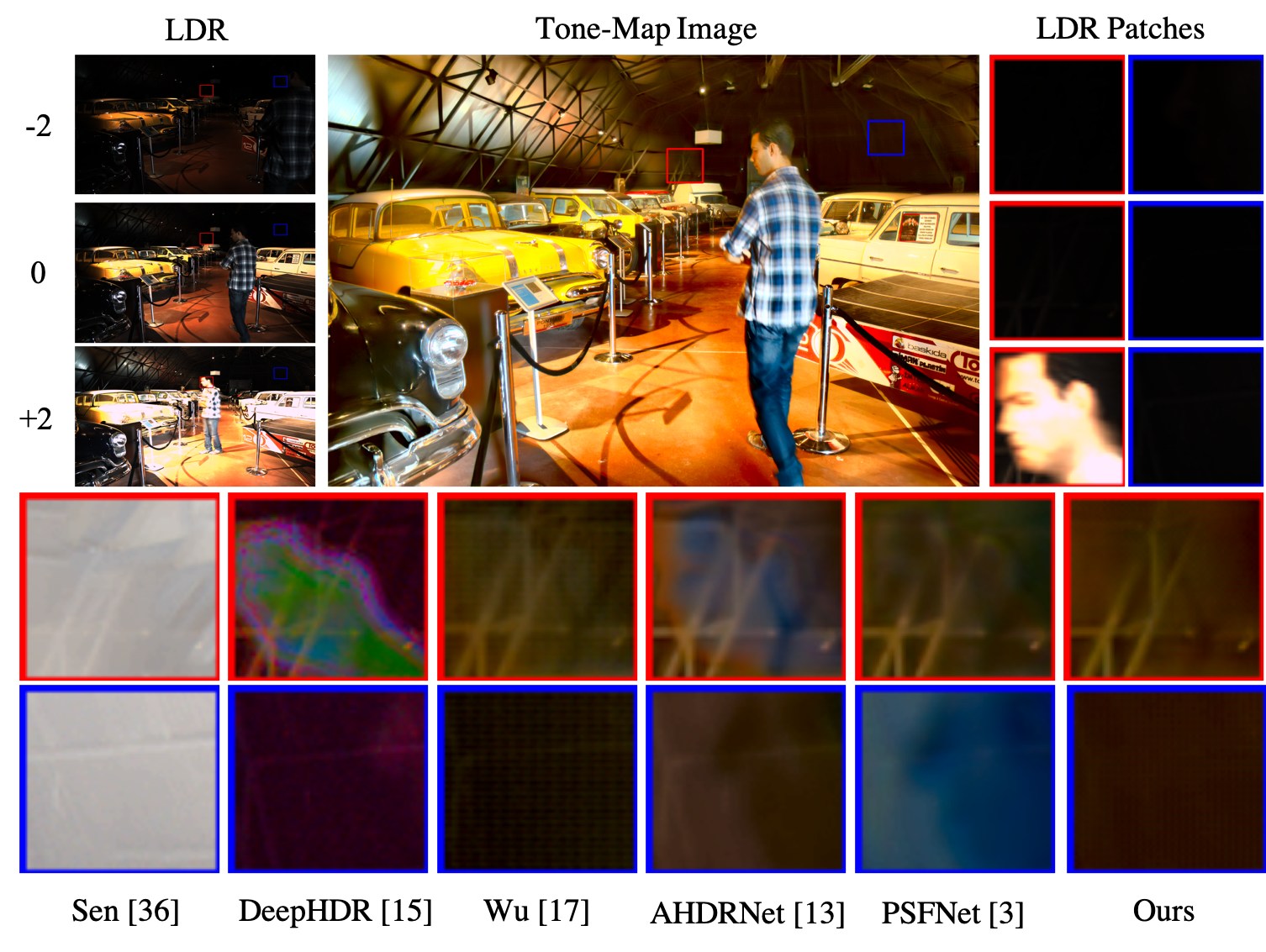}
}
\caption{Visual comparison of different HDR imaging methods. The image illustrated in (a) is selected from the Sen dataset, and the image shown in (b) is from the Tursun dataset.}
    \label{vis_2}
\end{figure*}

\subsection{Experiments on the Sen and Tursun Datasets}
In this experiment, we evaluated our proposed PFANet against other state-of-the-art HDR methods on two challenging datasets, i.e., the Sen and Tursun datasets. However, since these two datasets do not contain ground-truth images for quantitative comparison, we rely on visual results to compare the performance of different HDR imaging models. To ensure a fair comparison, we selected one image from each dataset and cropped two local regions marked by red and blue rectangles. These regions are then enlarged for better comparison, as shown in Fig.\,\ref{vis_2}. As observed, the HDR images generated by Sen and DeepHDR are noisy and contain unpleasant artifacts caused by object motion. Although AHDRNet and PSFNet are able to produce satisfactory results for the first dataset, severe ghosting artifacts are easily observed in underexposed regions containing object motion in the second dataset. Moreover, PSFNet fails to preserve object shapes, thus losing details. This implies that AHDRNet and PSFNet are vulnerable to changes in exposure conditions and have limited generalization capabilities. In contrast, our proposed model demonstrated excellent performance in reducing ghosting artifacts caused by object motion and preserving structural information in the generated HDR images. As a result, the HDR images generated by our model were cleaner, richer in detail, and less distorted, with the best perceptual quality.

\subsection{Model Analysis}

\subsubsection{Different numbers of sampled features} 
\begin{figure}[!t]
    \centering
    \includegraphics[width=0.6\linewidth]{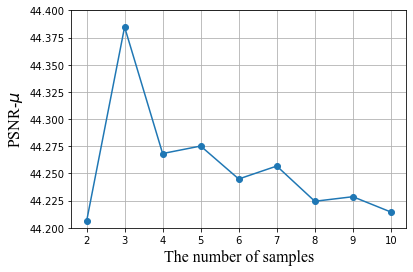}
    \caption{The PSNR-$\mu$ of the model with different numbers of sampled features.}
    \label{nums_samples}
\end{figure}

We have derived a novel method that utilizes similar features around unaligned features to implicitly align input LDR images in the feature space. To evaluate the effectiveness of our method, we conducted an experiment on the Kalantari dataset to investigate the impact of the number of sampled features on the performance. We vary the number of sampling features $N$ from 2 to 10 and evaluate the corresponding performance of the model. Fig.\,\ref{nums_samples} shows the PSNR-$\mu$ of the models with different numbers of sampling features. Our results demonstrate that the proposed method achieves the best performance when $N=3$. Interestingly, increasing the number of sampled features from 3 to 10 leads to a decrease in PSNR-$\mu$ by 0.15dB. This result is due to the inclusion of inaccurate sampled features in the aggregation process, which negatively affects performance.

\subsubsection{Modules in the Multi-scale Feature Alignment Sub-network} 
\begin{figure}[htbp]
    \centering
        \includegraphics[width=\linewidth]{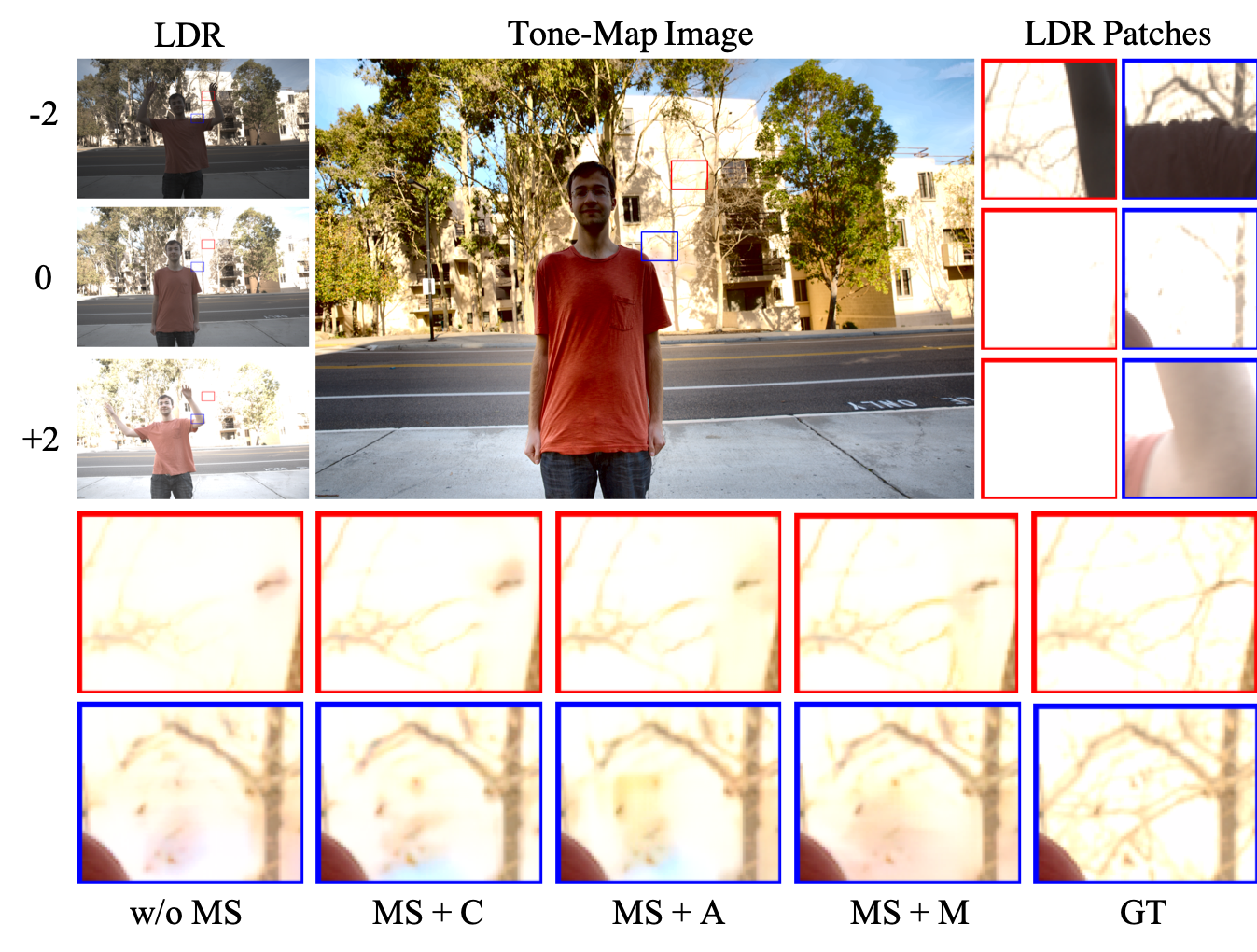}
    \caption{HDR images generated by our method (i.e.,\,MASNet) with different settings. ``w/o MS" represents the model without using the multi-scale structure. ``MS+C",``MS+A", and ``MS+M" represent that the model adopts concatenation, addition, and the proposed progressive masking method for fusion, respectively.}
    \label{MASNet_diff}
\end{figure}

The multi-scale feature alignment sub-network is a critical component in our proposed model, which implicitly aligns LDR images in a coarse-to-fine manner. To assess its performance, we conducted an experiment with different settings. Specifically, we evaluate our model performance with and without using the multi-scale structure, and compare the performance of different fusion methods across scales, including addition, concatenation, and our proposed inter-scale aggregation scheme. All other model configurations remain unchanged. The average PSNR-$\mu$, SSIM-$\mu$, and HDR-VDP-2 of our model with different settings on the Kalantari dataset are shown in Table \ref{aba1}. The best performance is highlighted in bold. The multi-scale structure based on our proposed inter-scale aggregation scheme yields the most significant improvement, increasing PSNR-$\mu$ by 0.55dB. The proposed aggregation scheme adaptively suppresses unnecessary pixels across scales, leading to superior performance compared to the addition and concatenation methods. Furthermore, Fig.\,\ref{MASNet_diff} demonstrates the images generated by our model with different settings for visual comparison. The model with our proposed multi-scale structure generates images with richer and more detailed information, such as tree branches, compared to the model without it. This result confirms that the multi-scale structure can effectively capture local details, resulting in better reconstruction. In contrast, models using addition and concatenation fusions produce images with color deviations, whereas our proposed inter-scale aggregation scheme significantly reduces these distortions and produces better images.

\begin{table}[htbp]
\centering
\caption{Average PSNR-$\mu$, SSIM-$\mu$, and HDR-VDP-2(VDP-2) of our proposed model (PFANet) with different settings on the Kalantari dataset. ``MS" represents the multi-scale structure. ``A", ``C", and ``M" denote the addition, concatenation, and our proposed masking method in inter-scale aggregation, respectively. ``\CheckmarkBold" represents the corresponding method used in PFANet, while ``\XSolidBrush" means that the method is not used.}
\begin{tabular}{|c|c|ccc|c|c|c|c|}
\hline
\multirow{2}{*}{Models} & \multicolumn{1}{c|}{\multirow{2}{*}{MS}} & \multicolumn{3}{c|}{Fusion}                                  & \multirow{2}{*}{PSNR-$\mu$} & \multirow{2}{*}{SSIM-$\mu$} & \multirow{2}{*}{VDP-2} \\ \cline{3-5}
                  & \multicolumn{1}{c|}{}                  & \multicolumn{1}{c|}{A} & \multicolumn{1}{c|}{C}       & M &                   &                 &  \\ \hline
PFANet   &     \XSolidBrush                                  &         \XSolidBrush               &        \XSolidBrush                    & \XSolidBrush    &     44.0933              &    0.9910             &  63.6499 \\ \cline{1-1} 
PFANet  &    \CheckmarkBold                                    &          \CheckmarkBold               &        \XSolidBrush      &  \XSolidBrush  &    44.2538               &      \textbf{0.9912}        & 64.3989    \\ \cline{1-1} 
PFANet  &    \CheckmarkBold                                      &          \XSolidBrush            &           \CheckmarkBold                   & \XSolidBrush &     44.1815              &     0.9911        & 64.1025    \\ \cline{1-1} 
PFANet  &    \CheckmarkBold                & \XSolidBrush  &\XSolidBrush  & \CheckmarkBold   &       \textbf{44.3847}            &   \textbf{0.9912}              & \textbf{64.5038} \\ \hline
\end{tabular}
\label{aba1}
\end{table}
\subsubsection{Study on the Dense Wavelet Sub-network} Our proposed dense wavelet sub-network utilizes DWT to decompose input features into several non-overlapping frequency subbands for better feature extraction and reconstruction. Additionally, we employ the group strategy and densely concatenate the outputs of the three Group WaveNets to fully utilize the information from intermediate layers. To evaluate the effectiveness of DWT in our method, we conducted an experiment comparing the performance of our model with and without DWT. In the model without DWT, feature extraction is simply performed over the spatial domain. All other configurations remain the same. Moreover, we study the impact of the number of groups. In our model, the total number of groups is set to 3, and we vary this number from 1 to 3. The average PSNR-$\mu$, SSIM-$\mu$, and HDR-VDP-2 of our model with different settings on the Kalantari dataset are tabulated in Table \ref{study_wavenet}. The results show that our model with DWT achieves PSNR 0.13dB higher than the model without DWT, indicating that extracting features in the wavelet domain is more effective, since different frequency subbands provide distinct information for HDR image reconstruction. Furthermore, our model performance improves significantly with an increase in the number of groups, since having more groups enhances the interaction between different layers.

\begin{table}[htbp]
\centering
\caption{Average PSNR-$\mu$, SSIM-$\mu$, and HDR-VDP-2 of the models with different settings on the Kalantari dataset. The best results for different settings are highlighted in bold.}
\begin{tabular}{|cc|c|c|c|}
\hline
\multicolumn{2}{|c|}{}                     & PSNR-$\mu$ & SSIM-$\mu$ & HDR-VDP-2  \\ \hline
\multicolumn{1}{|c|}{\multirow{2}{*}{Domain}}  & w DWT & \textbf{44.3847} & \textbf{0.9912} & \textbf{64.5038}  \\ \cline{2-5} 
\multicolumn{1}{|c|}{}                  & w/o DWT & 44.1566 & 0.9910 & 64.0267  \\ \hline \hline
\multicolumn{1}{|c|}{\multirow{3}{*}{Group}} & $G=1$ & 44.1203 & 0.9911 & 63.8058  \\ \cline{2-5} 
\multicolumn{1}{|c|}{}                  & $G=2$ & 44.2136 & 0.9910 & 64.3134 \\ \cline{2-5} 
\multicolumn{1}{|c|}{}                  & $G=3$ & \textbf{44.3847} & \textbf{0.9912} & \textbf{64.5038}  \\ \hline
\end{tabular}
\label{study_wavenet}
\end{table}

\subsubsection{Study on Different Types of Wavelets} The DWT-based subnetwork is a critical component for feature extraction in our model. To comprehensively study the impact of DWT, we investigate the performance of our model with different types of wavelets. Specifically, we compare the performance of our model with Haar wavelets, two symmetry wavelets, and two biorthogonal wavelets. Other configuration settings remain unchanged. The average PSNR-$\mu$ and SSIM-$\mu$ of our model with different types of wavelets on the Kalantari dataset are tabulated in Table \ref{study_wavelet}. As observed, different wavelets used in our model will significantly affect the performance. The performance of the model with symmetry wavelets is lower than that with Haar and biorthogonal wavelets. Compared with biorthogonal wavelets, the model with Haar wavelets can achieve slightly better performance because Haar wavelets are orthogonal. These results show that orthogonal and biorthogonal wavelets are more beneficial for reconstruction.

\begin{table}[htbp]
\centering
\caption{Average PSNR-$\mu$ and SSIM-$\mu$ of our model with different wavelets on the Kalantari dataset. The best results are highlighted in bold.}
\begin{tabular}{|c|c|cc|cc|}
\hline
\multirow{2}{*}{} & \multirow{2}{*}{Haar}    & \multicolumn{2}{c|}{Symmetry}    & \multicolumn{2}{c|}{Biorthogonal}    \\ \cline{3-6} 
                  &                   & \multicolumn{1}{c|}{sym2} & sym5  & \multicolumn{1}{l|}{bio1.1} & bio3.5  \\ \hline
PSNR-$\mu$        &   \textbf{44.38}  & \multicolumn{1}{c|}{44.29} & 44.08 & \multicolumn{1}{c|}{44.35} & 44.18 \\ \hline
SSIM-$\mu$        &   0.9912  & \multicolumn{1}{c|}{\textbf{0.9913}} & 0.9911 & \multicolumn{1}{c|}{0.9912} & 0.9912  \\ \hline
\end{tabular}
\label{study_wavelet}
\end{table}

\section{Conclusion}
High dynamic range (HDR) imaging in dynamic scenes remains a challenging problem in image processing and computer vision. In this paper, we propose a deep progressive feature aggregation network for HDR imaging in dynamic scenes, which implicitly aligns LDR images by sampling and aggregating similar features around unaligned pixels in different scale spaces. The aggregated features are then progressively fused in a coarse-to-fine manner, resulting in reduced ghosting artifacts caused by object motion. To further enhance the quality of the generated HDR images, we introduce a dense wavelet sub-network, which decomposes the input feature into non-overlapping frequency subbands for feature extraction. Different frequency subbands contain distinct information for reconstruction, enabling our method to effectively compensate for corrupted content in saturated regions. Experimental results demonstrate that our proposed method outperforms existing HDR methods, achieving state-of-the-art results with fewer distortions and more detailed content, leading to superior visual quality.

\bibliography{mybibfile}

\begin{thebibliography}{10}
\expandafter\ifx\csname url\endcsname\relax
  \def\url#1{\texttt{#1}}\fi
\expandafter\ifx\csname urlprefix\endcsname\relax\def\urlprefix{URL }\fi
\expandafter\ifx\csname href\endcsname\relax
  \def\href#1#2{#2} \def\path#1{#1}\fi

\bibitem{froehlich2014creating}
J.~Froehlich, S.~Grandinetti, B.~Eberhardt, S.~Walter, A.~Schilling,
  H.~Brendel, Creating cinematic wide gamut hdr-video for the evaluation of
  tone mapping operators and hdr-displays, in: Digital photography X, Vol.
  9023, International Society for Optics and Photonics, 2014, p. 90230X.

\bibitem{tocci2011versatile}
M.~D. Tocci, C.~Kiser, N.~Tocci, P.~Sen, A versatile hdr video production
  system, ACM Transactions on Graphics (TOG) 30~(4) (2011) 1--10.

\bibitem{yang2018adaptive}
K.-F. Yang, H.~Li, H.~Kuang, C.-Y. Li, Y.-J. Li, An adaptive method for image
  dynamic range adjustment, IEEE Transactions on Circuits and Systems for Video
  Technology 29~(3) (2018) 640--652.

\bibitem{liu2020single}
Y.-L. Liu, W.-S. Lai, Y.-S. Chen, Y.-L. Kao, M.-H. Yang, Y.-Y. Chuang, J.-B.
  Huang, Single-image hdr reconstruction by learning to reverse the camera
  pipeline, in: Proceedings of the IEEE Conference on Computer Vision and
  Pattern Recognition, 2020, pp. 1651--1660.

\bibitem{eilertsen2017hdr}
G.~Eilertsen, J.~Kronander, G.~Denes, R.~K. Mantiuk, J.~Unger, Hdr image
  reconstruction from a single exposure using deep cnns, ACM transactions on
  graphics (TOG) 36~(6) (2017) 1--15.

\bibitem{lee2020learning}
S.~Lee, S.~Y. Jo, G.~H. An, S.-J. Kang, Learning to generate multi-exposure
  stacks with cycle consistency for high dynamic range imaging, IEEE
  Transactions on Multimedia 23 (2020) 2561--2574.

\bibitem{tan2021deep}
X.~Tan, H.~Chen, K.~Xu, Y.~Jin, C.~Zhu, Deep sr-hdr: Joint learning of
  super-resolution and high dynamic range imaging for dynamic scenes, IEEE
  Transactions on Multimedia 25 (2023) 750--763.
\newblock \href {https://doi.org/10.1109/TMM.2021.3132165}
  {\path{doi:10.1109/TMM.2021.3132165}}.

\bibitem{fotiadou2019snapshot}
K.~Fotiadou, G.~Tsagkatakis, P.~Tsakalides, Snapshot high dynamic range imaging
  via sparse representations and feature learning, IEEE Transactions on
  Multimedia 22~(3) (2019) 688--703.

\bibitem{bogoni2000extending}
L.~Bogoni, Extending dynamic range of monochrome and color images through
  fusion, in: Proceedings of the International Conference on Pattern
  Recognition., Vol.~3, IEEE, 2000, pp. 7--12.

\bibitem{khan2006ghost}
E.~A. Khan, A.~O. Akyuz, E.~Reinhard, Ghost removal in high dynamic range
  images, in: Proceedings of the International Conference on Image Processing,
  IEEE, 2006, pp. 2005--2008.

\bibitem{jacobs2008automatic}
K.~Jacobs, C.~Loscos, G.~Ward, Automatic high-dynamic range image generation
  for dynamic scenes, IEEE Computer Graphics and Applications 28~(2) (2008)
  84--93.

\bibitem{reinhard2010high}
E.~Reinhard, W.~Heidrich, P.~Debevec, S.~Pattanaik, G.~Ward, K.~Myszkowski,
  High dynamic range imaging: acquisition, display, and image-based lighting,
  Morgan Kaufmann, 2010.

\bibitem{ye2021progressive}
Q.~Ye, J.~Xiao, K.-M. Lam, T.~Okatani, Progressive and selective fusion network
  for high dynamic range imaging, in: Proceedings of the 29th ACM International
  Conference on Multimedia, 2021, pp. 5290--5297.

\bibitem{gallo2009artifact}
O.~Gallo, N.~Gelfandz, W.-C. Chen, M.~Tico, K.~Pulli, Artifact-free high
  dynamic range imaging, in: 2009 IEEE International conference on
  computational photography (ICCP), IEEE, 2009, pp. 1--7.

\bibitem{grosch2006fast}
T.~Grosch, Fast and robust high dynamic range image generation with camera and
  object movement, Vision, Modeling and Visualization, RWTH Aachen (2006)
  277--284.

\bibitem{oh2014robust}
T.-H. Oh, J.-Y. Lee, Y.-W. Tai, I.~S. Kweon, Robust high dynamic range imaging
  by rank minimization, IEEE Transactions on Pattern Analysis and Machine
  Intelligence 37~(6) (2014) 1219--1232.

\bibitem{yan2017high}
Q.~Yan, J.~Sun, H.~Li, Y.~Zhu, Y.~Zhang, High dynamic range imaging by sparse
  representation, Neurocomputing 269 (2017) 160--169.

\bibitem{lee2014ghost}
C.~Lee, Y.~Li, V.~Monga, Ghost-free high dynamic range imaging via rank
  minimization, IEEE Signal Processing Letters 21~(9) (2014) 1045--1049.

\bibitem{yan2019attention}
Q.~Yan, D.~Gong, Q.~Shi, A.~v.~d. Hengel, C.~Shen, I.~Reid, Y.~Zhang,
  Attention-guided network for ghost-free high dynamic range imaging, in:
  Proceedings of the IEEE/CVF Conference on Computer Vision and Pattern
  Recognition, 2019, pp. 1751--1760.

\bibitem{pan2020multi}
Z.~Pan, M.~Yu, G.~Jiang, H.~Xu, Z.~Peng, F.~Chen, Multi-exposure high dynamic
  range imaging with informative content enhanced network, Neurocomputing 386
  (2020) 147--164.

\bibitem{yan2019robust}
Q.~Yan, Y.~Zhu, Y.~Zhang, Robust artifact-free high dynamic range imaging of
  dynamic scenes, Multimedia Tools and Applications 78~(9) (2019) 11487--11505.

\bibitem{kalantari2017deep}
N.~K. Kalantari, R.~Ramamoorthi, et~al., Deep high dynamic range imaging of
  dynamic scenes., ACM Transactions on Graphics (TOG) 36~(4) (2017) 144--1.

\bibitem{hafner2014simultaneous}
D.~Hafner, O.~Demetz, J.~Weickert, Simultaneous hdr and optic flow computation,
  in: Proceedings of the International Conference on Pattern Recognition, IEEE,
  2014, pp. 2065--2070.

\bibitem{wu2018deep}
S.~Wu, J.~Xu, Y.-W. Tai, C.-K. Tang, Deep high dynamic range imaging with large
  foreground motions, in: Proceedings of the European Conference on Computer
  Vision (ECCV), 2018, pp. 117--132.

\bibitem{zhang2011gradient}
W.~Zhang, W.-K. Cham, Gradient-directed multiexposure composition, IEEE
  Transactions on Image Processing 21~(4) (2011) 2318--2323.

\bibitem{lee2018multi}
S.-h. Lee, J.~S. Park, N.~I. Cho, A multi-exposure image fusion based on the
  adaptive weights reflecting the relative pixel intensity and global gradient,
  in: Proceedings of the IEEE International Conference on Image Processing
  (ICIP), IEEE, 2018, pp. 1737--1741.

\bibitem{liu2009beyond}
C.~Liu, et~al., Beyond pixels: exploring new representations and applications
  for motion analysis, Ph.D. thesis, Massachusetts Institute of Technology
  (2009).

\bibitem{sun2018pwc}
D.~Sun, X.~Yang, M.-Y. Liu, J.~Kautz, Pwc-net: Cnns for optical flow using
  pyramid, warping, and cost volume, in: Proceedings of the IEEE Conference on
  Computer Vision and Pattern Recognition (CVPR), 2018, pp. 8934--8943.

\bibitem{yu2016back}
J.~J. Yu, A.~W. Harley, K.~G. Derpanis, Back to basics: Unsupervised learning
  of optical flow via brightness constancy and motion smoothness, in:
  Proceedings of the European Conference on Computer Vision (ECCV), Springer,
  2016, pp. 3--10.

\bibitem{teed2020raft}
Z.~Teed, J.~Deng, Raft: Recurrent all-pairs field transforms for optical flow,
  in: Proceedings of the European Conference on Computer Vision, Springer,
  2020, pp. 402--419.

\bibitem{pu2020robust}
Z.~Pu, P.~Guo, M.~S. Asif, Z.~Ma, Robust high dynamic range (hdr) imaging with
  complex motion and parallax, in: Proceedings of the Asian Conference on
  Computer Vision, 2020.

\bibitem{zhu2019deformable}
X.~Zhu, H.~Hu, S.~Lin, J.~Dai, Deformable convnets v2: More deformable, better
  results, in: Proceedings of the IEEE/CVF Conference on Computer Vision and
  Pattern Recognition, 2019, pp. 9308--9316.

\bibitem{wang2019edvr}
X.~Wang, K.~C. Chan, K.~Yu, C.~Dong, C.~Change~Loy, Edvr: Video restoration
  with enhanced deformable convolutional networks, in: Proceedings of the
  IEEE/CVF Conference on Computer Vision and Pattern Recognition Workshops,
  2019, pp. 0--0.

\bibitem{liu2021adnet}
Z.~Liu, W.~Lin, X.~Li, Q.~Rao, T.~Jiang, M.~Han, H.~Fan, J.~Sun, S.~Liu, Adnet:
  Attention-guided deformable convolutional network for high dynamic range
  imaging, in: Proceedings of the IEEE/CVF Conference on Computer Vision and
  Pattern Recognition Workshop, 2021, pp. 463--470.

\bibitem{chan2021understanding}
K.~C. Chan, X.~Wang, K.~Yu, C.~Dong, C.~C. Loy, Understanding deformable
  alignment in video super-resolution, in: Proceedings of the AAAI Conference
  on Artificial Intelligence, Vol.~35, 2021, pp. 973--981.

\bibitem{raman2011reconstruction}
S.~Raman, S.~Chaudhuri, Reconstruction of high contrast images for dynamic
  scenes, The Visual Computer 27~(12) (2011) 1099--1114.

\bibitem{heo2010ghost}
Y.~S. Heo, K.~M. Lee, S.~U. Lee, Y.~Moon, J.~Cha, Ghost-free high dynamic range
  imaging, in: Proceedings of the Asian Conference on Computer Vision,
  Springer, 2010, pp. 486--500.

\bibitem{tomaszewska2007image}
A.~Tomaszewska, R.~Mantiuk, Image registration for multi-exposure high dynamic
  range image acquisition (2007).

\bibitem{zimmer2011freehand}
H.~Zimmer, A.~Bruhn, J.~Weickert, Freehand hdr imaging of moving scenes with
  simultaneous resolution enhancement, in: Computer Graphics Forum, Vol.~30,
  Wiley Online Library, 2011, pp. 405--414.

\bibitem{sen2012robust}
P.~Sen, N.~K. Kalantari, M.~Yaesoubi, S.~Darabi, D.~B. Goldman, E.~Shechtman,
  Robust patch-based hdr reconstruction of dynamic scenes., ACM Transactions on
  Graphics (TOG) 31~(6) (2012) 203--1.

\bibitem{yan2020deep}
Q.~Yan, L.~Zhang, Y.~Liu, Y.~Zhu, J.~Sun, Q.~Shi, Y.~Zhang, Deep hdr imaging
  via a non-local network, IEEE Transactions on Image Processing 29 (2020)
  4308--4322.

\bibitem{chen2022attention}
J.~Chen, Z.~Yang, T.~N. Chan, H.~Li, J.~Hou, L.-P. Chau, Attention-guided
  progressive neural texture fusion for high dynamic range image restoration,
  IEEE Transactions on Image Processing 31 (2022) 2661--2672.

\bibitem{niu2021hdr}
Y.~Niu, J.~Wu, W.~Liu, W.~Guo, R.~W. Lau, Hdr-gan: Hdr image reconstruction
  from multi-exposed ldr images with large motions, IEEE Transactions on Image
  Processing 30 (2021) 3885--3896.

\bibitem{mao2017least}
X.~Mao, Q.~Li, H.~Xie, R.~Y. Lau, Z.~Wang, S.~Paul~Smolley, Least squares
  generative adversarial networks, in: Proceedings of the IEEE International
  Conference on Computer Vision, 2017, pp. 2794--2802.

\bibitem{adler2018banach}
J.~Adler, S.~Lunz, Banach wasserstein gan, Proceedings of the Advances in
  Neural Information Processing Systems 31 (2018).

\bibitem{gulrajani2017improved}
I.~Gulrajani, F.~Ahmed, M.~Arjovsky, V.~Dumoulin, A.~C. Courville, Improved
  training of wasserstein gans, Proceedings of the Advances in neural
  information processing systems 30 (2017).

\bibitem{hu2018squeeze}
J.~Hu, L.~Shen, G.~Sun, Squeeze-and-excitation networks, in: Proceedings of the
  IEEE Conference on Computer Vision and Pattern Recognition, 2018, pp.
  7132--7141.

\end{thebibliography}

\end{document}